\documentclass[10pt,journal,compsoc]{IEEEtran}
\usepackage{amsmath,amsfonts}
\usepackage{algorithmic}
\usepackage{array}
\usepackage[caption=false,font=footnotesize,labelfont=sf,textfont=sf]{subfig}
\usepackage{textcomp}
\usepackage{stfloats}
\usepackage{url}
\usepackage{verbatim}
\usepackage{graphicx}
\usepackage[nocompress]{cite}
\usepackage[hidelinks]{hyperref}
\usepackage{multirow}
\usepackage{booktabs}
\usepackage{pifont}
\graphicspath{{./}}
\begin{document}

\title{Generate, Align, and Restore: MLLM-Empowered Diffusion for Generalizable Low-light Image Enhancement}

\author{Kaisheng~Pang,
Feng~Li,~\IEEEmembership{Member,~IEEE},
Chen Zhang,
Junjun Jiang,~\IEEEmembership{Senior Member,~IEEE},
Haoang Li, ~\IEEEmembership{Member,~IEEE},
Sam~Kwong,~\IEEEmembership{Fellow,~IEEE},
and Runmin~Cong,~\IEEEmembership{Senior Member,~IEEE}%
\IEEEcompsocitemizethanks{%
\IEEEcompsocthanksitem Kaisheng Pang and Haoang Li are with the Thrust of Robotics and Autonomous Systems, The Hong Kong University of Science and Technology (Guangzhou), Guangzhou 511453, China (e-mail: kaisheng.pang@connect.hkust-gz.edu.cn, haoangli@hkust-gz.edu.cn).
\IEEEcompsocthanksitem Feng Li is with the Key Laboratory of Knowledge Engineering with Big Data, Ministry of Education of China, Hefei 230601, China, and also with the Innovation School of Artificial Intelligence, Hefei University of Technology, Hefei 230601, China (e-mail: fengli@hfut.edu.cn).
\IEEEcompsocthanksitem Junjun Jiang is with the School of
Computer Science and Technology, Harbin Institute of Technology, Harbin 150001, China (e-mail: jiangjunjun@hit.edu.cn).
\IEEEcompsocthanksitem Sam Kwong is with the School of Data Science, Lingnan University, Hong Kong (e-mail: samkwong@ln.edu.hk).
\IEEEcompsocthanksitem Chen Zhang and Runmin Cong are with the School of Control Science and Engineering, Shandong University, Jinan 250061, China, and also with the Key Laboratory of Knowledge Engineering with Big Data, Ministry of Education of China, Hefei 230601, China. (e-mail: chen.zhang@mail.edu.cn and rmcong@sdu.edu.cn)
\IEEEcompsocthanksitem Corresponding author: Runmin Cong}%
}

\markboth{}%
{Pang \MakeLowercase{\textit{et al.}}: Generate, Align, and Restore}

\IEEEtitleabstractindextext{%
\begin{abstract}
Low-light image enhancement (LLIE) is essential to numerous vision applications operating under adverse illumination. Despite substantial advances, real-world nighttime enhancement remains challenging due to the scarcity of strictly aligned day-night pairs and complex degradations involving extreme darkness and heterogeneous artificial light sources, leading to fragile generalization. To overcome these challenges, we propose MrLIE (\textbf{M}LLM-empowered \textbf{reference} \textbf{L}ow-light \textbf{I}mage \textbf{E}nhancement), a novel reference-based unsupervised LLIE framework that synergizes the generative and reasoning capabilities of multimodal large language models (MLLMs) with conditional diffusion. MrLIE employs the MLLM as a high-fidelity reference synthesizer and diagnostic critic, iteratively generating, diagnosing, and correcting normal-light references directly from degraded nighttime observations. Based on this, our method governs the enhancement process through two complementary mechanisms, in which a confidence offset module rectifies spatial-geometric inconsistencies for reference reliability and a structure-anchored injection scheme harmonizes aligned reference cues, low-light structural constraints, and ControlNet additive anchoring during the denoising process. Extensive experiments on diverse benchmarks demonstrate that MrLIE achieves superior perceptual quality, natural appearance, and robust generalization.
\end{abstract}

}

\maketitle
\IEEEdisplaynontitleabstractindextext
\IEEEpeerreviewmaketitle

\IEEEraisesectionheading{\section{Introduction}\label{section_intro}}

\begin{figure*}[!t]
\centering
\includegraphics[width=\textwidth]{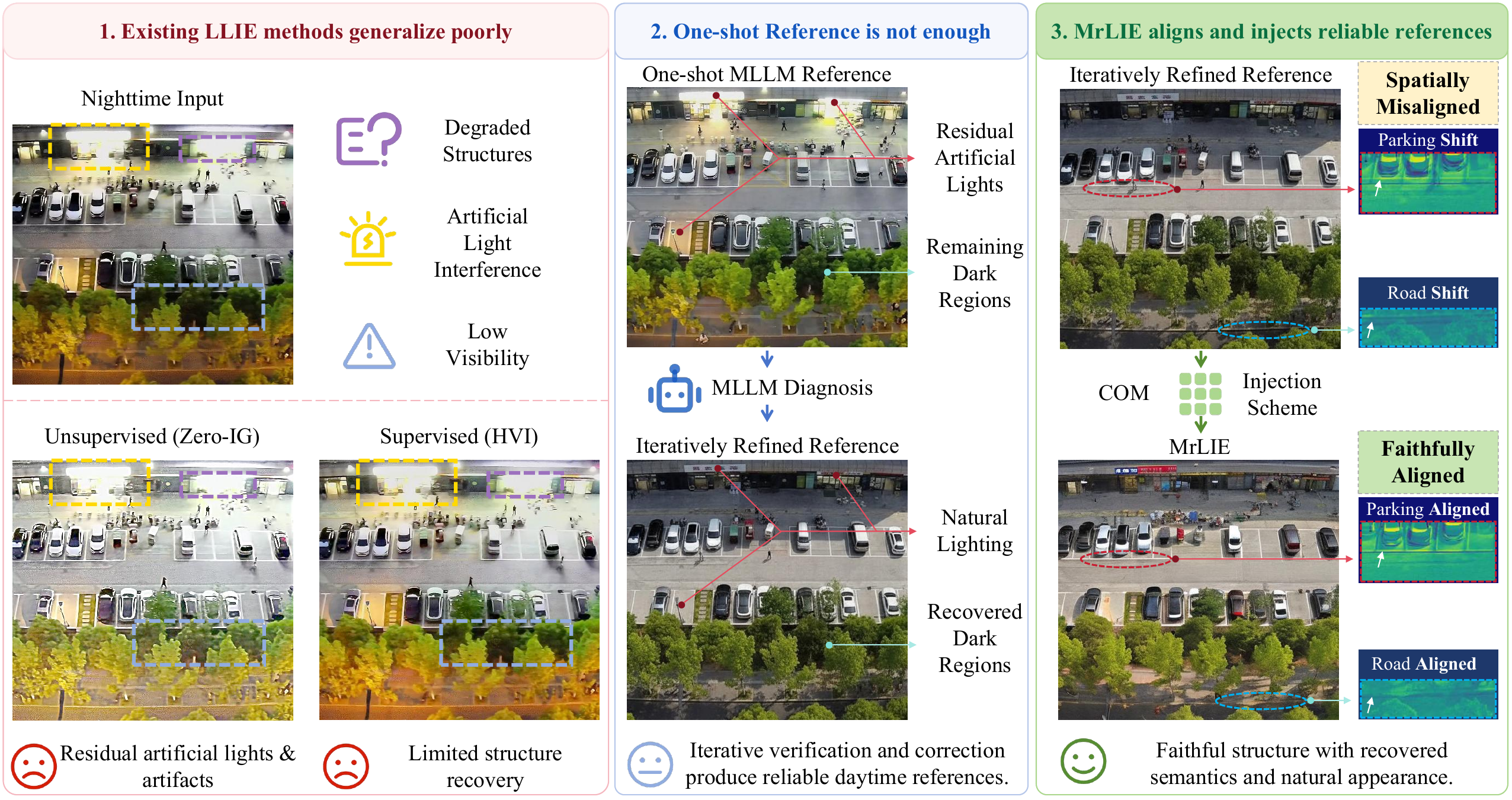}
\caption{
Motivation for MrLIE. Existing LLIE methods remain insufficient for real nighttime restoration, while one-shot daytime references often provide incomplete or spatially shifted guidance. MrLIE addresses these by iteratively constructing daytime references and aligning imperfect reference cues for faithful daytime restoration. The third-part visualization is obtained by subtracting the nighttime input from each corresponding image.
}
\label{fig1}
\end{figure*}
Low-light image enhancement (LLIE) seeks to recover degraded structures and chromatic information from images captured under adverse illumination~\cite{li2021low}, playing a critical role in downstream applications, such as surveillance~\cite{liu2021benchmarking,wang2021hla}, autonomous driving~\cite{sakaridis2021acdc, li2024light}, and computational photography~\cite{chen2018learning}.

In the literature, numerous LLIE methods have been proposed to improve image visibility and dramatically propelled with the continuous advent of deep learning techniques~\cite{chen2018learning,guo2020zero,zhang2021beyond,xu2022snr,cai2023retinexformer,wu2024towards,zhang2024noise,cong2026diffllface}. The dominant paradigm among these methods~\cite{wei2018deep,zhang2019kindling,yang2020fidelity,zhang2021beyond,wu2022uretinex,cai2023retinexformer,zhou2024glare} is heavily data-driven, dedicated to crafting powerful architectures relying on meticulously curated paired training data, encompassing long-exposure images, complex noise patterns, and non-uniform illumination, achieving impressive performance on standard benchmarks. However, these methods often generalize poorly to real-world nighttime scenes, where images suffer not only from underexposure but also from spatially varying illumination from heterogeneous artificial light sources (see Fig.~\ref{fig1}). Compounding this difficulty is the scarcity of strictly aligned day-night pairs in dynamic outdoor scenes, where simultaneous variations in illumination and degradation make pixel-level correspondence virtually unattainable.

To alleviate reliance on annotated pairings, later research has spurred growing interest in unpaired or zero-reference approaches. Unpaired methods leverage adversarial training to transfer high-quality style from unpaired normal-light images~\cite{jiang2021enlightengan,yang2023implicit,zhu2024unrolled,wu2025mutually}, yet remain highly sensitive to training data curation, as potential large discrepancies in scene content can cause unstable convergence and severe semantic misalignment. Zero-reference methods exploit physical or handcrafted priors derived from the low-light input itself~\cite{guo2020zero,li2021learning,ma2022toward,wang2024zero,shi2024zero,cho2026mr}. Though they offer broader practical applicability, the absence of explicit normal-light guidance can limit their ability to capture illumination distributions faithfully, resulting in poor naturalness and obvious visual artifacts.

Recently, multimodal large language models (MLLMs) have demonstrated exceptional semantic understanding and reasoning capabilities, even on degraded visual inputs~\cite{malik2025robust,tang2026robust,wei2026iamagent}, making them effective priors for fine-grained control in visual generation. This suggests that we can leverage MLLMs to provide valuable scene-specific visual knowledge unseen in low-light corruptions without the necessity for accurately aligned image pairs. To this end, we propose MrLIE (\textbf{MLLM-empowered} \textbf{reference} \textbf{Low-light} \textbf{Image} \textbf{Enhancement}), which harnesses MLLMs as the high-fidelity reference synthesizer and discerning critic, furnishing semantically consistent guidance to facilitate unsupervised LLIE. Nevertheless, although foundational MLLMs excel at visual generation, semantic consistency alone cannot strictly guarantee perceptual naturalness or physical plausibility (Fig.~\ref{fig1} (middle)). To alleviate this problem, MrLIE orchestrates a systematic reasoning schedule that integrates proactive visual synthesis with generative diagnostics in a closed-loop process, enabling autonomous correction of unreasonable flaws to improve reliability.

Furthermore, given that potential spatial shifts in generative references can induce texture hallucinations in misaligned local regions (Fig.~\ref{fig1} (right)), MrLIE regulates the enhancement process through two synergistic mechanisms. First, we devise the confidence offset module (COM) that predicts learnable offset fields and confidence maps to align imperfect references with the target scene geometry, ensuring rigorous structural consistency. Second, we develop a structure-anchored injection (SAI) scheme that harmonizes the reference, low-light structural constraint, and ControlNet additive anchoring into a consolidated information flow for controllable generation. The main contributions of this work are as follows:
\begin{itemize}
    \item We propose MrLIE, a novel reference-based unsupervised LLIE approach that connects the generative and reasoning capabilities of MLLMs with the conditional LLIE diffusion framework. Experiments on diverse benchmarks demonstrate its superior robustness and generalizability.

    \item We introduce a systematic MLLM-based generation pipeline that couples instruction-guided synthesis with iterative self-correction, enabling high-fidelity normal-light reference generation.

    \item We develop two complementary mechanisms, where the COM mitigates spatial misalignment through coordinate rectification and confidence mapping, and the SAI synergizes multi-source guidance with input structural anchors for plausible texture recovery.
\end{itemize}

\section{Related Works}\label{section_relatedworks}
\subsection{Low-Light Image Enhancement}
LLIE has advanced rapidly in recent years, driven substantially by the development of deep learning~\cite{lore2017llnet,chen2018learning,wei2018deep,zhang2019kindling,zhang2021beyond,xu2022snr,wu2022uretinex,wang2023ultra,cai2023retinexformer,yan2025hvi}. Supervised deep models learn photometric correction and denoising directly from paired images~\cite{lore2017llnet,chen2018learning}. LLNet~\cite{lore2017llnet} uses a stacked sparse denoising autoencoder to brighten and denoise synthetically degraded images. Xu~\emph{et al.}~\cite{xu2022snr} incorporate signal-to-noise ratio (SNR) priors to avoid noise influence in very low-SNR regions. LLFormer~\cite{wang2023ultra} introduces axis-based attention for efficient ultra-high-resolution enhancement. Another prominent line builds upon Retinex theory by explicitly modeling illumination decomposition~\cite{wei2018deep,zhang2019kindling,wu2022uretinex,cai2023retinexformer}. For instance,  RetinexNet~\cite{wei2018deep} separates reflectance and illumination. RetinexFormer~\cite{cai2023retinexformer} combines a one-stage Retinex formulation with illumination-guided non-local modeling. Diffusion models (DMs) have also been introduced into LLIE to leverage their strong generative capabilities~\cite{yi2023diff,hou2023global,chan2024anlightendiff,lan2025efficient}. Diff-Retinex~\cite{yi2023diff} learns conditional diffusion processes for normal-light reflectance and illumination. AnlightenDiff~\cite{chan2024anlightendiff} anchors iterative denoising to the low-light observation to improve perceptual quality while preserving image content. Despite remarkable performance, these methods remain trained with strong supervision on aligned low-light and normal-light image pairs, which are costly and impractical to collect in dynamic scenes. Moreover, these datasets synthesized under limited conditions cannot fully represent complex real-world degradations, leading to poor generalization to unseen environments.

\begin{figure*}[t]
\centering
\includegraphics[width=\textwidth]{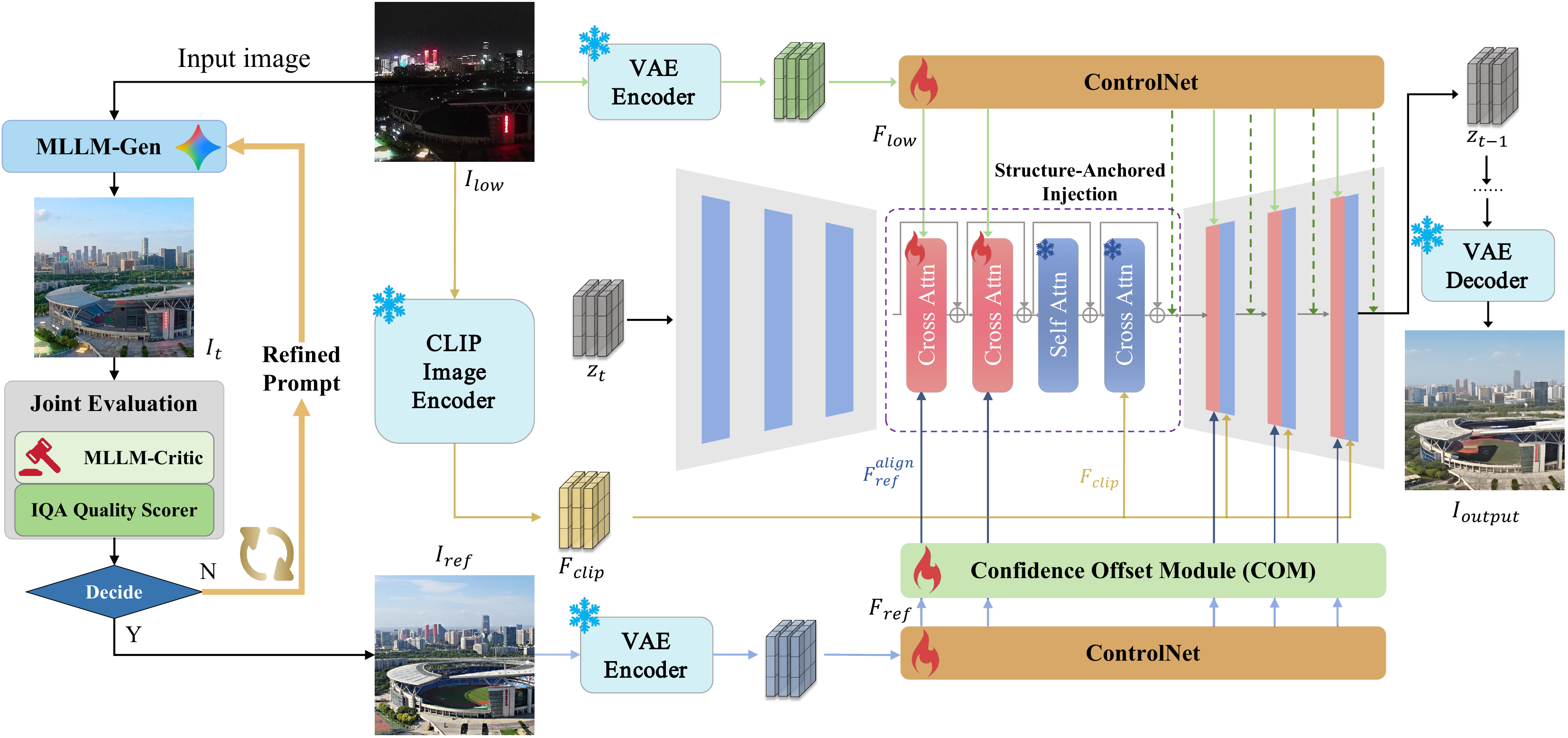}
\caption{Overall framework of the proposed MrLIE. It first obtains a high-quality normal-light reference through iterative MLLM-guided generation, then performs aligned-reference diffusion restoration by jointly injecting low-light structure, aligned reference features, and diffusion control signals.}\label{fig-pipeline}
\end{figure*}

\subsection{Unsupervised Low-light Image Enhancement}
To reduce the dependence on aligned training pairs, unsupervised LLIE methods exploit independently collected normal-light images as distributional supervision~\cite{jiang2021enlightengan,yang2023implicit,yang2025difflle}. EnlightenGAN~\cite{jiang2021enlightengan} learns to transfer normal-light characteristics through an attention-guided adversarial framework. NeRCo~\cite{yang2023implicit} uses implicit neural representations and cooperative constraints to preserve content during enhancement. On the other hand, zero-reference methods avoid strict supervision by exploiting priors or constraints derived from the input image itself~\cite{guo2020zero,li2021learning,ma2022toward,wang2024zero,shi2024zero,cho2026mr}. Zero-DCE and Zero-DCE++~\cite{guo2020zero,li2021learning} learn pixel-wise adjustment curves using carefully designed no-reference losses. SCI~\cite{ma2022toward} introduces self-calibrated illumination learning for efficient enhancement. Zero-IG~\cite{shi2024zero} jointly performs illumination-guided denoising and adaptive enhancement in a zero-shot manner. MR. Illuminate~\cite{cho2026mr} further exploits pretrained diffusion priors for zero-shot dark-to-light image conversion. These methods offer broad applicability but depend strongly on physical assumptions or external priors.

\subsection{Multimodal Large Language Models for Vision}
MLLMs connect visual encoders with large language models (LLMs), extending language-based reasoning and instruction following to visual content~\cite{li2023blip,liu2023visual,chen2024internvl}. BLIP-2~\cite{li2023blip} bridges frozen image encoders and language models through a lightweight query transformer. LLaVA~\cite{liu2023visual} develops visual instruction tuning for general-purpose visual understanding. Motivated by the impressive advances of MLLMs, recent works employ MLLMs to guide visual synthesis and editing~\cite{fu2024guiding,fang2025puma}. MGIE~\cite{fu2024guiding} translates user instructions into expressive guidance for image editing. PUMA~\cite{fang2025puma} unifies multi-granular visual features to support diverse generation. This is also increasingly explored in image restoration~\cite{zhang2023adding,wang2024exploiting,chen2024restoreagent,lin2024diffbir,yu2024scaling,jiang2026multi}, which constructs MLLM-based restoration agents to introduce explicit reasoning into the restoration pipeline, effectively processing both degradation perception and suitable restoration lines. RestoreAgent~\cite{chen2024restoreagent} constructs a restoration system that adaptively identifies the degradation types and builds a restoration sequence. MAIR~\cite{jiang2026multi} coordinates multiple agents to reason about degradation removal and image quality improvement. However, they mainly use MLLMs to interpret instructions or organize restoration tools and typically assume that reliable guidance is readily available.

\begin{figure*}[t]
\centering
\includegraphics[width=\textwidth]{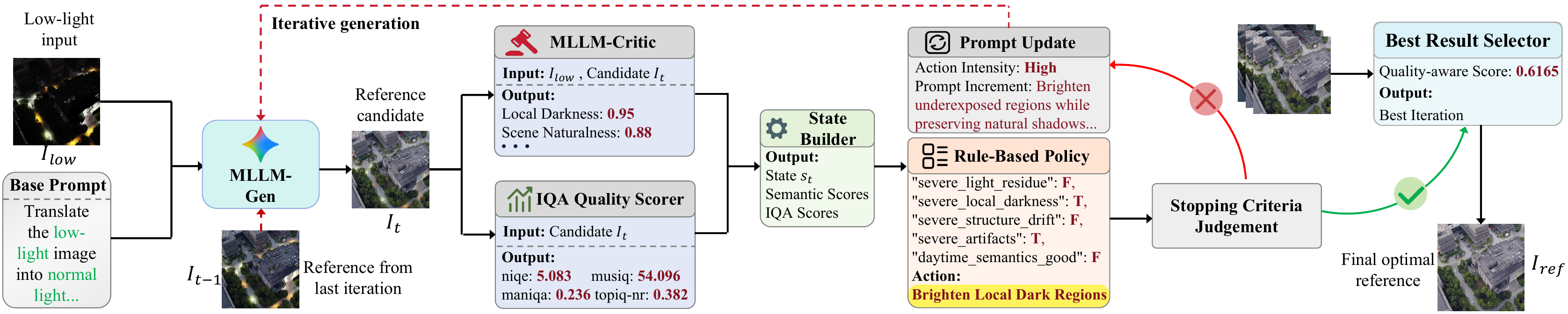}
\caption{Diagnosis-driven daytime reference generation. MLLM-Gen produces candidates from the nighttime input, while MLLM-Critic and no-reference image quality assessment jointly diagnose residual degradations and guide prompt refinement until a reliable reference is selected.}\label{fig-gene}
\end{figure*}

\section{Methodology}\label{section_method}
\subsection{Overview}
\label{sec:overview_pipeline}
As illustrated in Fig.~\ref{fig-pipeline}, the proposed MrLIE aims to restore a high-quality normal-light image $I_{output}$ from a low-light input $I_{low}$ through a reference-guided diffusion pipeline. The overall framework consists of two stages. We first generate a semantically enriched normal-light reference image $I_{ref}$ from $I_{low}$, and then use both the low-light observation and the generated reference $I_{ref}$ to guide the subsequent diffusion restoration process. MrLIE is built upon three key components. First, we introduce an MLLM-guided iterative generation process to obtain a reliable reference with richer semantic cues, as detailed in \textbf{Sec.~\ref{sec:iterative_ref_generation}}. Second, considering that the generated reference is not strictly aligned with the input, we design a confidence offset module (COM) to explicitly correct local spatial misalignment, as described in \textbf{Sec.~\ref{sec:sga}}. Third, based on the aligned reference, we develop a structure-anchored injection scheme, where reference guidance, low-light structural constraint, and conventional ControlNet additive anchoring are jointly incorporated into the diffusion backbone for robust restoration, as introduced in \textbf{Sec.~\ref{sec:triple_injection}}. Our method is reference-based and unsupervised to achieve realistic restoration. The training objective of the whole framework is given in \textbf{Sec.~\ref{sec:loss}}.

\subsection{MLLM-based Reference Generation}
\label{sec:iterative_ref_generation}
We propose to generate a reliable reference with sufficient information from low-light observations by exploiting recent advances in MLLMs. However, as illustrated in Fig.~\ref{fig1}, while MLLMs possess vast knowledge priors for visual synthesis, simple one-step generation often yields unsatisfying results characterized by residual luminance artifacts and regions of physical implausibility. To address this, we develop a reference-guided iterative generation framework, MLLM-Gen, which integrates semantic diagnosis, perceptual quality assessment, and action-guided prompt refinement. Starting with the input $I_{low}$, MLLM-Gen first generates an initial candidate under a base textual prompt and then progressively refines it until the desired semantics and visual quality are achieved, formulated as
\begin{equation}
I_0 = G(I_{low}, p_0), \qquad I_t = G(I_{low}, p_t, {I}_{t-1}), \; t \geq 1,
\end{equation}
where $p_t$ is the refinement prompt at iteration $t$. $I_t$ is the generated candidate. ${I}_{t-1}$ is fed back as an auxiliary reference for iterative refinement.

Then, each candidate is evaluated from two complementary perspectives. First, we employ MLLM-Critic to compare $I_{low}$ with the current candidate $I_t$ and assess whether plausible semantics have been established. The semantic feedback at iteration $t$ is defined as
\begin{equation}
c_t = C(I_{low}, I_t) = [D_t, L_t, K_t, F_t, A_t, N_t],
\end{equation}
where $D_t$, $L_t$, $K_t$, $F_t$, $A_t$, and $N_t$ denote semantic consistency, artificial-light residue, local darkness, structural fidelity, artifact level, and scene naturalness, respectively. To improve reliability, we impose explicit rules on the critic output: \emph{if the returned tags indicate remaining artificial illumination, the residue score is prevented from being overly low, and the candidate will not be considered eligible for termination}.

Second, we introduce a no-reference image quality assessment (IQA) scorer that aggregates normalized NIQE~\cite{mittal2012making}, MUSIQ~\cite{ke2021musiq}, MANIQA~\cite{yang2022maniqa}, and TOPIQ-NR~\cite{chen2024topiq} to provide a compact quality estimation. The two branches are jointly used to determine whether further refinement is required, since semantic plausibility alone may still coexist with visual artifacts, while images with acceptable perceptual quality may still preserve obvious low-light cues.

After that, the two types of assessments are fed into a structured decision state for action selection. Additionally, this state also introduces threshold-based indicators that explicitly capture severe failure modes, defined as
\begin{equation}
s_t = B(c_t, Q_t, H_{t-1}), \qquad a_t = \pi(s_t, H_{t-1}),
\end{equation}
where $s_t$ denotes the decision state and $Q_t$ denotes the comprehensive image assessment at iteration $t$. $H_{t-1}$ denotes the refinement history before step $t$. $B(\cdot)$ denotes the state construction process. $a_t$ denotes the selected refinement action. $\pi(\cdot)$ denotes the rule-based policy that determines the next action, including suppressing artificial lights, enhancing the normal-light appearance, brightening local dark regions, reducing artifacts, and improving naturalness.

Here, actions are selected in a fixed priority order, where dominant semantic failures are addressed first to ensure interpretability and stability. The selected actions are converted into the next refinement prompt by combining them with the base prompt to encourage the preservation of improved parts while avoiding the reintroduction of low-light components or distorted geometry until the optimal reference is obtained. Moreover, each action is associated with a discrete strength level that controls the emphasis of the refinement instruction in the prompt. To avoid ineffective repetition, repeated actions are downgraded when recent improvements become marginal.

\subsection{Confidence Offset Module}
\label{sec:sga}
\begin{figure}[t]
\centering
\includegraphics[width=\linewidth]{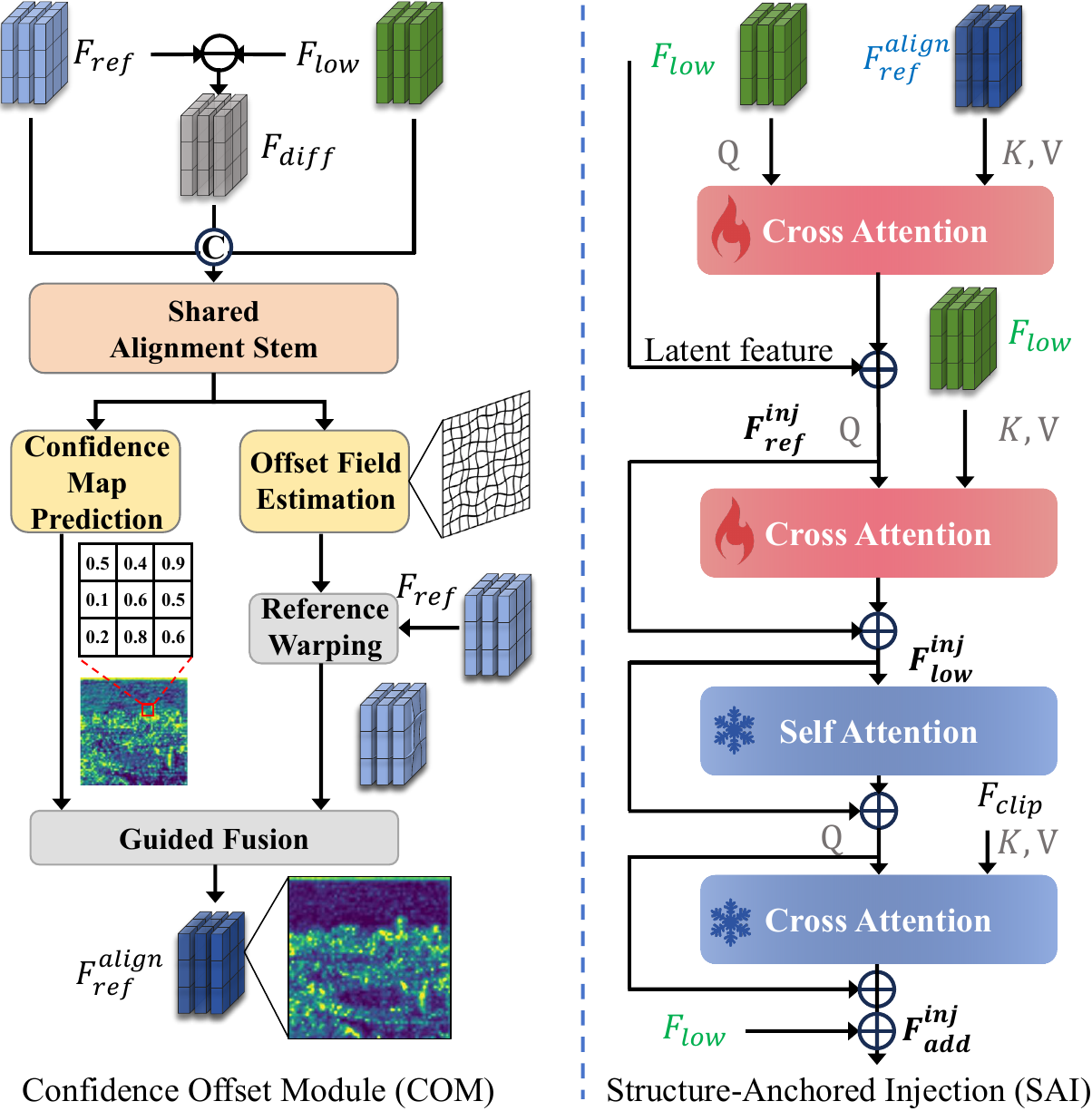}
\caption{Overview of the proposed Confidence Offset Module (COM) and Structure-Anchored Injection (SAI). COM aligns the generated reference features with the low-light features, while SAI injects reference and structural cues into the diffusion process for faithful enhancement.}\label{fig-consai}
\end{figure}

Before integrating the reference information into the subsequent diffusion process, the COM is devised to explicitly correct the spatial misalignment between the low-light input and the reference guidance. As shown in Fig.~\ref{fig-consai}, given the low-light feature $F_{low}$ and the reference feature $F_{ref}$, COM estimates an offset field and a spatial confidence map from their joint representation and then uses them to obtain an aligned reference feature $F_{ref}^{align}$.

\textbf{Offset Field Estimation}. COM first computes the absolute discrepancy $F_{diff}$ between $F_{low}$ and $F_{ref}$:
\begin{equation}
F_{diff}=|F_{low}-F_{ref}|.
\end{equation}
$F_{diff}$ highlights regions that differ in local structure and illumination. These three components are concatenated and fed into a lightweight residual network to produce a discrepancy-aware feature $F_s$. COM predicts a dense two-dimensional offset field from $F_s$ to model the local spatial correspondence
\begin{equation}
\Delta p = \mathcal{H}_{offset}(F_s),
\end{equation}
where $\mathcal{H}_{\text{offset}}(\cdot)$ denotes a learnable offset prediction head, and $\Delta p\in\mathbb{R}^{2\times H\times W}$ denotes the pixel-space displacement field along the horizontal and vertical directions. $\mathcal{H}_{\text{offset}}(\cdot)$ provides a location-dependent correction signal, allowing each position to adaptively retrieve auxiliary information.

For a spatial location $p=(x,y)$ on the target feature grid, COM computes its corresponding sampling location in the reference feature:
\begin{equation}
p_s = p-\Delta p(p).
\end{equation}
The sampling coordinate is normalized for grid sampling
\begin{equation}
G(p)=
\left(
\frac{2p_s^x}{W-1}-1,\,
\frac{2p_s^y}{H-1}-1
\right).
\end{equation}
We use bilinear interpolation to sample the reference feature from continuous spatial coordinates, resulting in the warped feature $\widetilde{F}_{ref}$ as
\begin{equation}
\widetilde{F}_{ref}=\mathrm{GridSample}(F_{ref},G).
\end{equation}

In this way, $\widetilde{F}_{ref}$ is spatially recalibrated and provides an explicit remedy for local misalignments, ensuring that the generative guidance remains robust even in the presence of significant structural shifts or texture displacements.

\textbf{Confidence Map Prediction.} In parallel with offset estimation, COM predicts a confidence map $M\in[0,1]^{1\times H\times W}$ from the same shared feature $F_s$:
\begin{equation}
M=\sigma(\mathcal{H}_{conf}(F_s)),
\end{equation}
where $\sigma(\cdot)$ denotes the sigmoid function. $M$ reveals the reliability of the aligned reference information at each spatial location. A large value indicates that the warped reference feature should contribute more strongly, while a smaller value suppresses potentially unreliable or incorrectly aligned reference information. Finally, the original reference feature $F_{ref}$ and the warped one $\widetilde{F}_{ref}$ are fused under the guidance of $M$:
\begin{equation}
F_{ref}^{align}=F_{ref}+M\odot(\widetilde{F}_{ref}-F_{ref}),
\end{equation}
where $\odot$ denotes element-wise multiplication. This formulation allows COM to adaptively inject aligned reference information while preserving the original reference information in uncertain regions.

\subsection{Structure-Anchored Injection}
\label{sec:triple_injection}
After COM, we propose a structure-anchored injection scheme that incorporates the complementary information from low-light input and reference guidance within the diffusion process. In the first cross-attention block (Fig.~\ref{fig-consai} (right)), $F_{low}$ serves as the query and the aligned reference control $F_{ref}^{align}$ provides the key and value, enabling the model to selectively absorb appearance details and semantic cues from the reference. In the second cross-attention block, the output of the preceding interaction is used as the query, while $F_{low}$ provides key and value, constraining the representation enhanced by the reference back to the structural content of low-light input. Besides the attention interactions, we also introduce a ControlNet-based structural injection branch, where the low-light control feature is injected through an additive pathway with zero convolution to provide explicit structural anchoring.


\subsection{Optimization Objective}
\label{sec:loss}

The proposed restoration model is optimized in the latent diffusion space. Since paired normal-light ground truth is unavailable, we use the generated reference as a pseudo target for training. The pseudo target is encoded into a latent representation $z_0$, while the low-light input and the aligned reference features are used as structural and appearance conditions. At each iteration, the timestep $t$ and Gaussian noise $\epsilon\sim\mathcal{N}(0,I)$ are sampled to obtain the noisy latent
\begin{equation}
z_t=\sqrt{\bar{\alpha}_t}z_0+\sqrt{1-\bar{\alpha}_t}\epsilon,
\end{equation}
where $\bar{\alpha}_t$ denotes the cumulative noise schedule. Let $\Phi=\{F_{low}, F_{ref}^{align}\}$ denote the conditional features formed by the low-light feature and the aligned reference feature produced by COM. Following the $\epsilon$-prediction formulation~\cite{saharia2022image}, the denoising objective is written as
\begin{equation}
\mathcal{L}_{\mathrm{diff}}
=
\mathbb{E}_{z_0,t,\epsilon}
\left[
\left\|
\epsilon_\theta(z_t,t,\Phi)-\epsilon
\right\|_2^2
\right].
\end{equation}

In addition, two lightweight regularization terms are introduced to stabilize COM across the multi-scale control features. We denote the offset field and confidence map at the $s$-th scale as $\Delta p^{(s)}$ and $M^{(s)}$, respectively. The offset smoothness loss $\mathcal{L}_{\mathrm{off}}$ is used to encourage locally coherent deformation
\begin{equation}
\mathcal{L}_{\mathrm{off}}
=
\frac{1}{S}\sum_{s=1}^{S}
\left(
\|\nabla_x \Delta p^{(s)}\|_1+\|\nabla_y \Delta p^{(s)}\|_1
\right),
\end{equation}
where $S$ denotes the control scale number. The confidence regularization loss $\mathcal{L}_{\mathrm{conf}}$ keeps the average reference gate around a neutral target $\tau$, preventing the model from collapsing to either unconditional reference injection or complete suppression
\begin{equation}
\mathcal{L}_{\mathrm{conf}}
=
\frac{1}{S}\sum_{s=1}^{S}
\left(
\mathrm{Mean}(M^{(s)})-\tau
\right)^2 .
\end{equation}
In our experiments, we use a neutral target of $\tau=0.5$ to avoid biased reference gating. Therefore, the overall training objective is
\begin{equation}
\label{eq:full_loss}
\mathcal{L}
=
\mathcal{L}_{\mathrm{diff}}
+
\lambda_{\mathrm{off}}\mathcal{L}_{\mathrm{off}}
+
\lambda_{\mathrm{conf}}\mathcal{L}_{\mathrm{conf}} .
\end{equation}

\section{Experiments}\label{sec-exp}
\begin{table*}[!tbp]
\caption{Quantitative comparison on DICM and NPE. Best and second-best results are highlighted in bold and underlined, respectively.}
\label{tab:dicm_npe_comparison}
\centering
\footnotesize
\setlength{\tabcolsep}{4.5pt}
\begingroup
\fontsize{8}{10}\selectfont
\renewcommand{\arraystretch}{1}
\setlength{\tabcolsep}{2.1pt}
\begin{tabular*}{\textwidth}{@{\extracolsep{\fill}}cll*{12}{c}@{}}
\toprule
\multirow{2}{*}{Type} & \multirow{2}{*}{Method} & \multirow{2}{*}{Venue} & \multicolumn{6}{c}{DICM} & \multicolumn{6}{c}{NPE} \\
\cmidrule(lr){4-9}\cmidrule(lr){10-15}
  &  &  & NIQE$\downarrow$ & MUSIQ$\uparrow$ & BRISQUE$\downarrow$ & PI$\downarrow$ & CLIP-IQA$\uparrow$ & LIQE$\uparrow$ & NIQE$\downarrow$ & MUSIQ$\uparrow$ & BRISQUE$\downarrow$ & PI$\downarrow$ & CLIP-IQA$\uparrow$ & LIQE$\uparrow$ \\
\midrule
\multirow{9}{*}{\textbf{S}} & LLFormer & AAAI'23 & 3.695 & 59.314 & 15.720 & 3.128 & 0.502 & 3.129 & 3.804 & 67.587 & 13.563 & 2.696 & 0.516 & 3.894 \\
  & RetinexFormer & ICCV'23 & 4.365 & 56.604 & 28.089 & 3.902 & 0.503 & 2.971 & 4.010 & 64.493 & 19.134 & 3.280 & 0.458 & 3.384 \\
  & DFFN & TGRS'24 & 3.746 & 55.429 & 17.242 & 3.214 & 0.397 & 2.314 & \underline{3.570} & 63.620 & 14.709 & 2.841 & 0.316 & 3.019 \\
  & HVI & CVPR'25 & 3.545 & 57.976 & 22.226 & \underline{3.080} & 0.588 & 3.143 & 3.879 & 67.560 & 18.153 & 2.923 & 0.546 & 3.698 \\
  & LBM & ICCV'25 & 4.813 & 60.868 & 26.486 & 3.953 & 0.560 & 3.637 & 4.485 & 68.028 & 18.070 & 3.214 & 0.576 & \underline{4.569} \\
  & GT-mean-loss & ICCV'25 & \underline{3.460} & 58.845 & \underline{15.080} & 3.167 & 0.470 & 2.755 & 3.882 & 63.553 & 18.580 & 2.834 & 0.340 & 3.149 \\
  & DATPRL-IR & ICLR'26 & 3.828 & 59.977 & 22.129 & 3.288 & 0.545 & 3.292 & 4.132 & 64.631 & 15.284 & 2.976 & \underline{0.626} & 3.815 \\
  & TPCNet & ECCV'26 & 3.726 & 58.723 & 15.512 & 3.089 & 0.549 & 3.198 & 4.040 & 67.927 & 12.735 & 2.856 & 0.488 & 4.019 \\
  & SLAIR & ECCV'26 & 4.604 & 59.471 & 25.964 & 3.701 & 0.590 & 3.353 & 4.077 & 65.037 & 14.532 & 2.855 & 0.615 & 3.836 \\
\midrule
\multirow{10}{*}{\textbf{U}} & Zero-DCE & CVPR'20 & 3.733 & 58.857 & 23.080 & 3.109 & 0.514 & 3.182 & 3.929 & 68.965 & 15.742 & 2.858 & 0.442 & 3.890 \\
  & RUAS & CVPR'21 & 7.053 & 49.433 & 46.522 & 5.552 & 0.285 & 1.498 & 7.065 & 55.864 & 49.643 & 5.194 & 0.204 & 1.646 \\
  & Zero-DCE++ & TPAMI'22 & 3.834 & 57.921 & 20.938 & 3.339 & 0.510 & 3.137 & 4.024 & 67.624 & 13.086 & 2.968 & 0.434 & 3.898 \\
  & NeRCo & ICCV'23 & 4.234 & 55.324 & 27.621 & 3.834 & 0.324 & 2.280 & 4.698 & 57.932 & 21.013 & 3.868 & 0.253 & 2.309 \\
  & PairLIE & CVPR'23 & 4.267 & 58.416 & 29.000 & 3.392 & 0.396 & 2.514 & 4.210 & 64.548 & 25.830 & 3.098 & 0.314 & 2.737 \\
  & Zero-IG & CVPR'24 & 5.840 & 51.953 & 48.611 & 4.522 & 0.369 & 1.514 & 7.742 & 60.633 & 67.624 & 5.061 & 0.243 & 1.807 \\
  & AGLLDiff & AAAI'25 & 4.891 & 60.320 & 37.530 & 4.199 & 0.433 & 3.326 & 4.762 & 66.898 & 27.978 & 3.681 & 0.386 & 3.572 \\
  & GEFU & ICCV'25 & 4.055 & \underline{62.198} & 18.958 & 3.285 & \underline{0.613} & \underline{3.738} & 3.690 & \underline{69.329} & \underline{12.117} & \underline{2.689} & 0.553 & 4.268 \\
  & ZeroIDIR & CVPR'26 & 3.761 & 59.883 & 21.080 & 3.237 & 0.532 & 3.447 & 3.903 & 69.038 & 17.244 & 2.900 & 0.492 & 4.145 \\
  & \textbf{MrLIE} & -- & \textbf{3.406} & \textbf{69.541} & \textbf{8.534} & \textbf{2.900} & \textbf{0.705} & \textbf{4.244} & \textbf{3.547} & \textbf{71.064} & \textbf{9.604} & \textbf{2.634} & \textbf{0.679} & \textbf{4.829} \\
\bottomrule
\end{tabular*}
\endgroup
\vspace{-0.3cm}
\end{table*}

\begin{table*}[!tbp]
\caption{Quantitative comparison on LIME and MEF. Best and second-best results are highlighted in bold and underlined, respectively.}
\label{tab:lime_mef_comparison}
\centering
\footnotesize
\setlength{\tabcolsep}{4.5pt}
\begingroup
\fontsize{8.5}{10.5}\selectfont
\renewcommand{\arraystretch}{1}
\setlength{\tabcolsep}{2.5pt}
\begin{tabular*}{\textwidth}{@{\extracolsep{\fill}}cl*{12}{c}@{}}
\toprule
\multirow{2}{*}{Type} & \multirow{2}{*}{Method} & \multicolumn{6}{c}{LIME} & \multicolumn{6}{c}{MEF} \\
\cmidrule(lr){3-8}\cmidrule(lr){9-14}
  &  & NIQE$\downarrow$ & MUSIQ$\uparrow$ & BRISQUE$\downarrow$ & PI$\downarrow$ & CLIP-IQA$\uparrow$ & LIQE$\uparrow$ & NIQE$\downarrow$ & MUSIQ$\uparrow$ & BRISQUE$\downarrow$ & PI$\downarrow$ & CLIP-IQA$\uparrow$ & LIQE$\uparrow$ \\
\midrule
\multirow{9}{*}{S} & LLFormer & 4.153 & 58.442 & 15.460 & 3.184 & 0.441 & 2.690 & 3.650 & 59.257 & 16.815 & 3.079 & 0.444 & 3.037 \\
  & RetinexFormer & 4.667 & 60.480 & 25.917 & 3.753 & 0.436 & 2.899 & 4.272 & 58.789 & 29.772 & 3.616 & 0.448 & 3.223 \\
  & DFFN & 3.923 & 60.924 & 16.384 & 2.821 & 0.410 & 2.876 & 3.755 & 62.567 & 22.944 & 2.958 & 0.445 & 3.290 \\
  & HVI & 4.163 & 62.378 & 20.714 & 3.172 & 0.461 & 3.308 & 3.697 & 66.736 & 24.588 & 2.854 & 0.474 & 3.980 \\
  & LBM & 4.637 & \underline{65.983} & 21.902 & 3.330 & 0.535 & \underline{3.798} & 4.435 & 64.865 & 24.377 & 3.228 & 0.541 & \underline{4.068} \\
  & GT-mean-loss & 3.791 & 61.410 & \underline{13.370} & 2.852 & 0.463 & 3.148 & 3.490 & 62.096 & 15.298 & 2.767 & 0.502 & 3.405 \\
  & DATPRL-IR & 4.148 & 61.014 & 19.591 & 3.121 & 0.506 & 3.119 & 3.811 & 65.653 & 24.340 & 2.947 & 0.485 & 3.799 \\
  & TPCNet & 3.807 & 60.020 & 14.480 & 2.937 & 0.490 & 2.978 & 3.277 & 62.590 & \textbf{10.987} & 2.506 & 0.525 & 3.458 \\
  & SLAIR & 4.750 & 57.904 & 21.168 & 3.550 & 0.552 & 2.753 & 4.398 & 58.916 & 21.920 & 3.360 & 0.551 & 3.052 \\
\midrule
\multirow{10}{*}{U} & Zero-DCE & \underline{3.771} & 60.735 & 18.483 & \underline{2.760} & 0.455 & 3.063 & 3.283 & 64.923 & 16.630 & \underline{2.434} & 0.520 & 3.786 \\
  & RUAS & 5.372 & 55.426 & 29.259 & 3.923 & 0.324 & 1.861 & 5.427 & 55.580 & 34.263 & 4.068 & 0.352 & 2.252 \\
  & Zero-DCE++ & 3.972 & 60.489 & 16.960 & 2.953 & 0.470 & 3.141 & 3.394 & 63.514 & 13.755 & 2.535 & 0.524 & 3.720 \\
  & NeRCo & 4.334 & 55.831 & 28.646 & 3.751 & 0.295 & 2.237 & 4.926 & 51.854 & 32.278 & 4.550 & 0.258 & 2.127 \\
  & PairLIE & 4.513 & 59.409 & 25.237 & 3.262 & 0.380 & 2.244 & 4.170 & 62.650 & 29.539 & 3.095 & 0.405 & 2.837 \\
  & Zero-IG & 4.529 & 56.466 & 28.953 & 3.188 & 0.398 & 1.989 & 4.266 & 62.201 & 25.418 & 2.964 & 0.535 & 2.809 \\
  & AGLLDiff & 5.032 & 65.653 & 31.767 & 3.862 & 0.436 & 3.269 & 5.016 & 65.772 & 37.739 & 4.068 & 0.371 & 3.690 \\
  & GEFU & 4.108 & 63.836 & 14.783 & 2.869 & \underline{0.597} & 3.539 & 3.691 & \underline{66.767} & 15.824 & 2.550 & \underline{0.659} & 3.799 \\
  & ZeroIDIR & 3.826 & 62.841 & 18.702 & 2.803 & 0.499 & 3.216 & \underline{3.232} & 63.825 & 15.470 & 2.497 & 0.538 & 3.734 \\
  & \textbf{MrLIE} & \textbf{3.309} & \textbf{70.031} & \textbf{10.187} & \textbf{2.617} & \textbf{0.683} & \textbf{4.351} & \textbf{3.189} & \textbf{68.891} & \underline{11.682} & \textbf{2.349} & \textbf{0.698} & \textbf{4.354} \\
\bottomrule
\end{tabular*}
\endgroup
\end{table*}

\begin{table*}[!tbp]
\caption{Quantitative comparison on U3D and the iSAID-dark validation benchmark. Unsupervised methods are trained on U3D where applicable and directly evaluated on iSAID-dark without fine-tuning.}
\label{tab:quantitative_comparison}
\label{tab:isaid_comparison}
\centering
\footnotesize
\setlength{\tabcolsep}{3pt}
\begingroup
\fontsize{7.5}{10}\selectfont
\renewcommand{\arraystretch}{1}
\setlength{\tabcolsep}{1.3pt}
\begin{tabular*}{\textwidth}{@{\extracolsep{\fill}}clcccccc@{\hspace{1em}}ccccccccc@{}}
\toprule
\multirow{2}{*}{Type} & \multirow{2}{*}{Method} & \multicolumn{6}{c}{U3D} & \multicolumn{9}{c}{iSAID-dark} \\
\cmidrule(lr){3-8}\cmidrule(lr){9-17}
  &  & NIQE$\downarrow$ & MUSIQ$\uparrow$ & BRISQUE$\downarrow$ & PI$\downarrow$ & CLIP-IQA$\uparrow$ & LIQE$\uparrow$ & PSNR$\uparrow$ & SSIM$\uparrow$ & LPIPS$\downarrow$ & NIQE$\downarrow$ & MUSIQ$\uparrow$ & BRISQUE$\downarrow$ & PI$\downarrow$ & CLIP-IQA$\uparrow$ & LIQE$\uparrow$ \\
\midrule
\multirow{9}{*}{\textbf{S}} & LLFormer & 4.938 & 38.948 & 24.583 & 3.227 & 0.253 & 1.341 & 14.744 & 0.290 & 0.988 & 6.690 & 39.601 & 46.091 & 4.420 & 0.236 & 1.016 \\
  & RetinexFormer & 3.788 & 38.995 & 20.401 & 2.850 & 0.246 & 1.356 & \underline{17.521} & 0.353 & 0.705 & 6.931 & 40.405 & 34.031 & 4.417 & 0.275 & 1.063 \\
  & DFFN & 5.374 & 37.678 & 32.314 & 3.556 & 0.264 & 1.270 & 14.997 & 0.265 & 0.878 & 9.676 & 38.719 & 59.826 & 6.076 & 0.276 & 1.054 \\
  & HVI & 5.370 & 37.777 & 37.914 & 3.338 & 0.285 & 1.462 & 13.301 & 0.202 & 1.065 & 10.715 & 37.895 & 64.101 & 6.523 & 0.294 & 1.044 \\
  & LBM & 5.449 & \underline{58.362} & 30.503 & 3.510 & 0.555 & 2.871 & 16.394 & 0.406 & 0.476 & 9.396 & 55.322 & 48.769 & 5.697 & 0.468 & \underline{2.446} \\
  & GT-mean-loss & 5.531 & 37.239 & 28.590 & 3.529 & 0.349 & 1.479 & 13.226 & 0.241 & 0.773 & 12.563 & 37.202 & 62.951 & 7.508 & 0.310 & 1.046 \\
  & DATPRL-IR & 6.392 & 38.089 & 33.458 & 4.102 & 0.333 & 1.462 & 10.956 & 0.423 & \textbf{0.374} & 5.787 & 53.386 & 42.860 & 4.334 & 0.372 & 1.915 \\
  & TPCNet & 5.551 & 37.395 & 36.656 & 3.426 & 0.297 & 1.442 & 14.257 & 0.195 & 0.931 & 12.246 & 36.615 & 59.910 & 7.035 & 0.247 & 1.058 \\
  & SLAIR & 4.895 & 38.035 & 18.914 & 3.339 & 0.327 & 1.336 & 10.430 & 0.354 & 0.460 & 6.435 & 49.987 & 34.127 & 4.716 & 0.426 & 1.890 \\
\midrule
\multirow{10}{*}{U} & Zero-DCE & 6.571 & 43.790 & 46.796 & 3.992 & 0.379 & 1.475 & 9.050 & 0.156 & 1.123 & 16.412 & 38.233 & 68.658 & 9.526 & 0.487 & 1.021 \\
  & RUAS & 6.137 & 35.401 & 40.663 & 4.134 & 0.231 & 1.073 & 9.960 & 0.162 & 0.899 & 10.616 & 35.776 & 56.881 & 6.533 & 0.417 & 1.011 \\
  & Zero-DCE++ & 6.301 & 36.798 & 44.147 & 3.854 & 0.356 & 1.417 & 9.015 & 0.159 & 1.120 & 16.309 & 36.403 & 69.036 & 9.426 & 0.481 & 1.019 \\
  & NeRCo & \underline{3.376} & 55.001 & \underline{11.247} & \underline{2.480} & 0.436 & 2.356 & 16.568 & 0.565 & 0.469 & \underline{3.980} & \underline{56.953} & \underline{11.120} & \underline{2.781} & 0.550 & 2.397 \\
  & PairLIE & 5.288 & 46.224 & 29.478 & 3.381 & 0.269 & 1.540 & 15.175 & 0.337 & 0.845 & 18.532 & 51.724 & 53.127 & 10.264 & 0.325 & 1.054 \\
  & Zero-IG & 6.530 & 48.464 & 50.759 & 4.031 & 0.347 & 1.364 & 10.703 & 0.139 & 1.108 & 17.440 & 41.637 & 75.866 & 10.050 & 0.516 & 1.025 \\
  & AGLLDiff & 7.832 & 46.166 & 52.812 & 4.874 & 0.278 & 1.443 & 13.879 & \underline{0.593} & \underline{0.429} & 5.135 & 45.903 & 38.900 & 3.685 & 0.208 & 1.148 \\
  & GEFU & 4.367 & 56.470 & 31.723 & 2.825 & \underline{0.564} & \textbf{3.619} & 14.585 & 0.318 & 0.858 & 9.705 & 55.575 & 53.901 & 5.709 & \underline{0.595} & 2.184 \\
  & ZeroIDIR & 5.382 & 42.359 & 36.215 & 3.385 & 0.311 & 1.738 & \textbf{17.698} & \textbf{0.611} & 0.433 & 5.085 & 51.839 & 27.733 & 3.387 & 0.211 & 1.472 \\
  & \textbf{MrLIE} & \textbf{2.884} & \textbf{66.710} & \textbf{5.368} & \textbf{2.110} & \textbf{0.637} & \underline{3.528} & 15.490 & 0.437 & 0.455 & \textbf{3.347} & \textbf{63.152} & \textbf{7.857} & \textbf{2.397} & \textbf{0.670} & \textbf{3.225} \\
\bottomrule
\end{tabular*}
\endgroup
\end{table*}

\subsection{Experimental Settings}
\textbf{Datasets}. We evaluate our method on six benchmarks comprising four standard real-world LLIE datasets, DICM~\cite{lee2013contrast}, NPE~\cite{wang2013naturalness}, LIME~\cite{guo2016lime}, and MEF~\cite{ma2015perceptual}, with only low-light images available, an unpaired low-altitude unmanned aerial vehicle (UAV) dataset U3D~\cite{lu2025unsupervised}, and a paired low-light remote sensing dataset iSAID-dark~\cite{yao2024spatial}.

\textbf{DICM} contains 69 real low-contrast photographs from diverse indoor and outdoor scenes. \textbf{NPE} comprises 8 representative images with non-uniform illumination captured across cloudy, dusk, and nighttime conditions. \textbf{LIME} includes 10 real low-light images with non-uniform illumination. \textbf{MEF} contains 17 underexposed images from multi-exposure sequences with strong brightness variations. \textbf{U3D} is an unsupervised ultra-high-resolution UAV low-light enhancement benchmark designed for real aerial imaging scenarios. It contains 1,000 real nocturnal UAV images and 1,000 unpaired daytime images covering diverse urban, residential, and natural scenes from aerial viewpoints. Following the official 8:1:1 split, we use 800 for training and 100 for testing. For \textbf{iSAID-dark}, we use its validation split, which contains 66 paired aerial images at $500\times500$ resolution from iSAID~\cite{waqas2019isaid} synthesized under diverse exposure levels and noise conditions.

\textbf{Evaluation Metrics}. We adopt NIQE~\cite{mittal2012making}, BRISQUE~\cite{mittal2012no}, PI~\cite{blau20182018}, MUSIQ~\cite{ke2021musiq}, CLIP-IQA~\cite{wang2023exploring}, and LIQE~\cite{zhang2023blind} to evaluate quantitative performance, where lower NIQE, BRISQUE, and PI scores and higher MUSIQ, CLIP-IQA, and LIQE scores indicate better perceptual quality. For the paired iSAID-dark benchmark, we additionally report full-reference PSNR, SSIM~\cite{wang2004image}, and LPIPS~\cite{zhang2018unreasonable}.

\textbf{Implementation Details}.
In MrLIE, MLLM-Gen is implemented with Gemini-3.1, while MLLM-Critic uses Gemini-2.5. The generated reference images are produced at $1344\times768$ resolution, which are then cropped into $512\times512$ patches, keeping the same spatial resolution as the low-light input. We train MrLIE on LOL for cross-dataset evaluation on DICM, NPE, LIME, and MEF. Given the domain gap between ground-view and aerial scenes, we train a separate model solely on U3D for evaluation on both U3D and iSAID-dark. MrLIE is trained with a batch size of 16 and a learning rate of $5.0\times10^{-5}$ for 8K iterations using the AdamW optimizer. We adopt the Stable Diffusion v2.1 pretrained model~\cite{rombach2022high} for diffusion restoration. The full framework is implemented in PyTorch on a single NVIDIA H100 GPU. The regularization weights in Eq.~\eqref{eq:full_loss} are set to $\lambda_{\mathrm{off}}=10^{-3}$ and $\lambda_{\mathrm{conf}}=5\times10^{-4}$.

\begin{figure*}[!tbp]
\centering
\includegraphics[width=\textwidth]{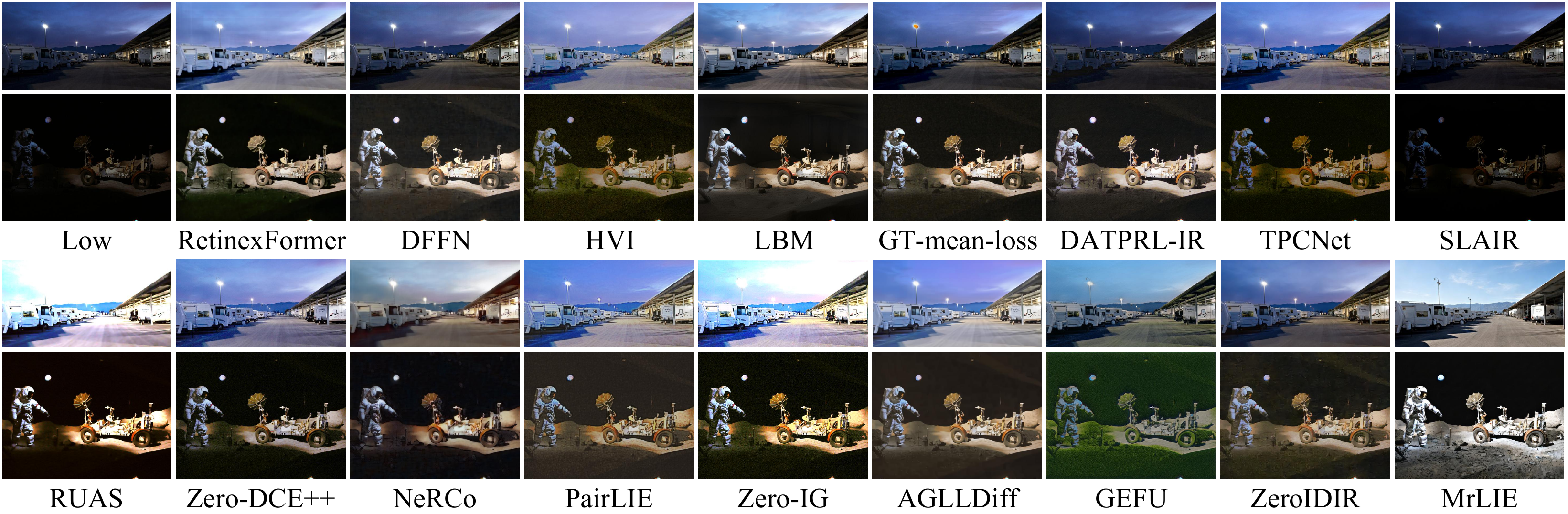}
\caption{Qualitative generalization results on DICM and NPE. MrLIE produces natural exposure and color while avoiding excessive enhancement, color distortion, and structural drift under real-world image domain shifts.}\label{fig-dicmnpe}
\end{figure*}

\begin{figure*}[!tbp]
\centering
\includegraphics[width=\textwidth]{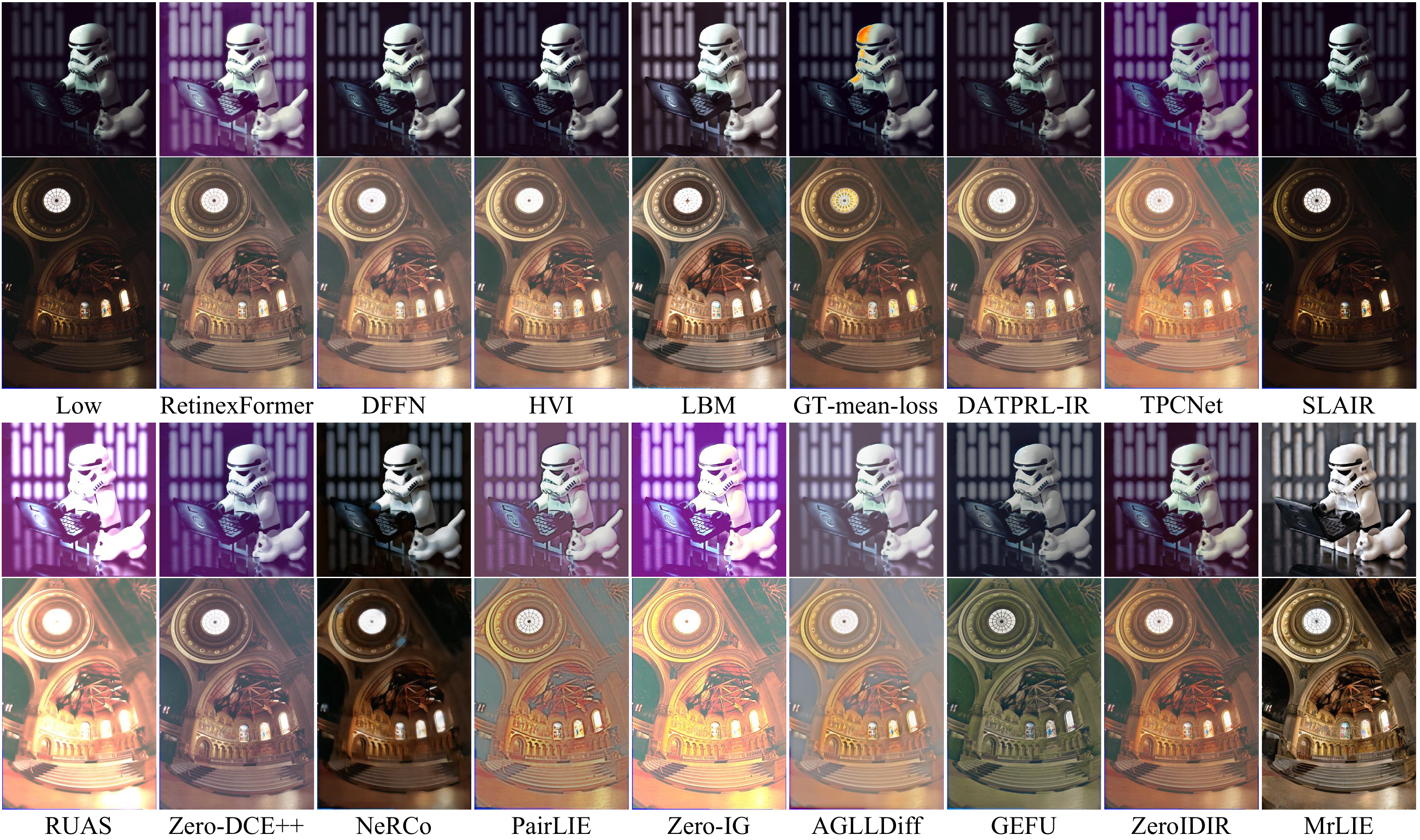}
\caption{Qualitative generalization results on LIME and MEF. The examples highlight differences in color fidelity and exposure balance, particularly around the helmet, ceiling details, and bright window regions.}
\label{fig-limenpe}
\end{figure*}

\begin{figure*}[!t]
\centering
\includegraphics[width=\textwidth]{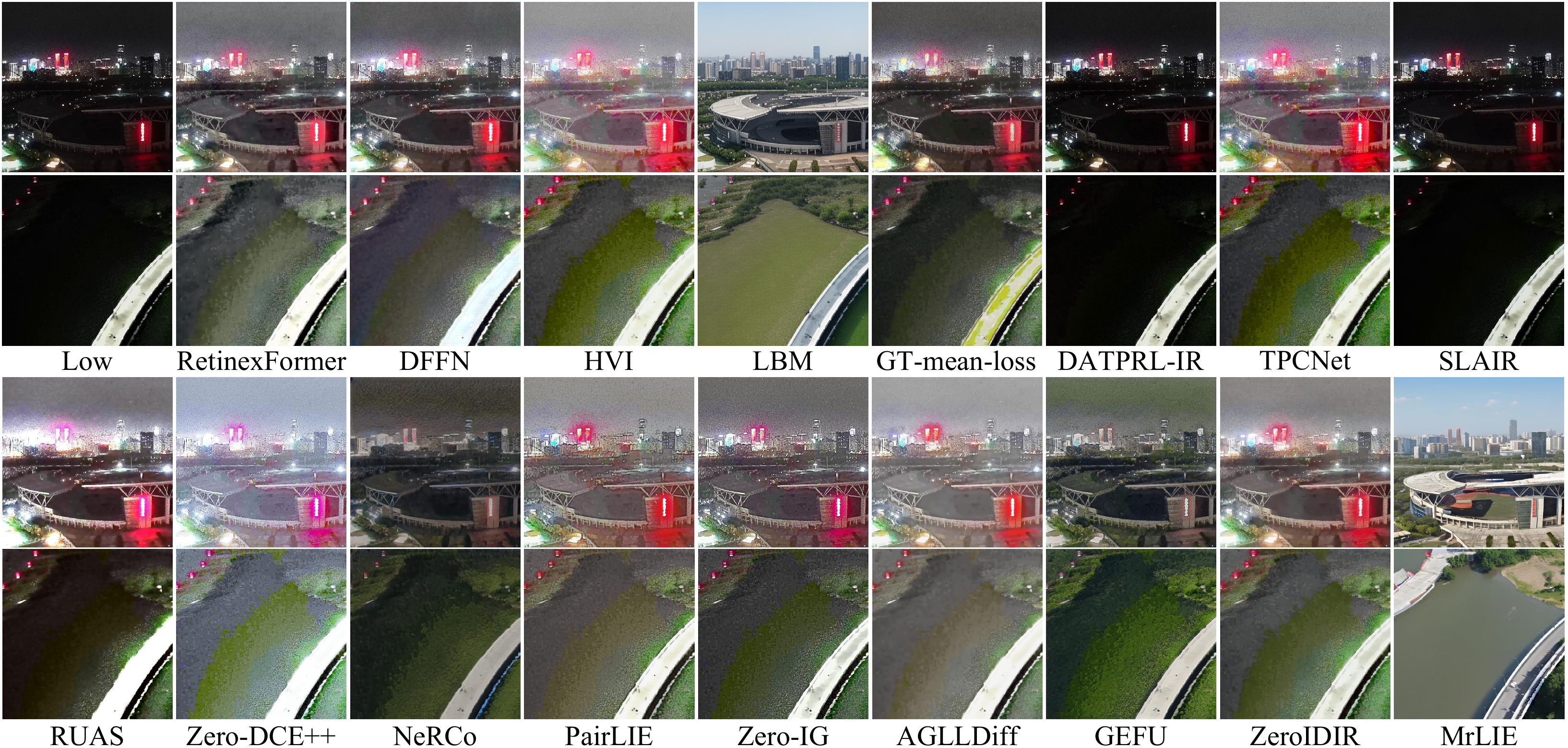}
\caption{Qualitative comparison on U3D nighttime aerial scenes. MrLIE reduces artificial-light residues and recovers daytime appearance while retaining the stadium outline and riverbank geometry.}\label{fig-u3d}
\par\medskip
\centering
\includegraphics[width=\textwidth]{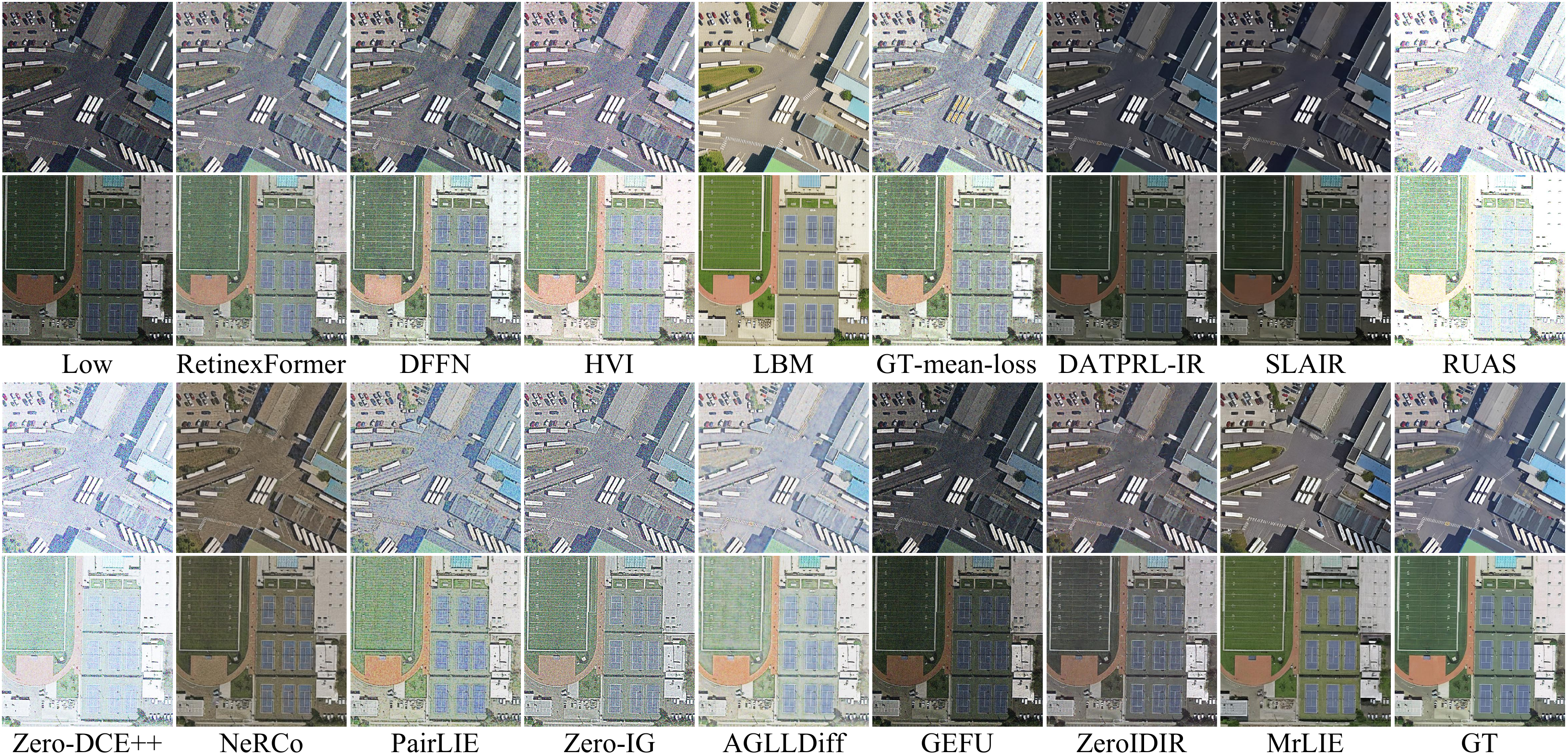}
\caption{Qualitative results on the iSAID-dark dataset. MrLIE achieves balanced illumination, clear aerial structures, and natural color appearance.}
\label{fig-iSAID}
\end{figure*}

\begin{figure}[!htbp]
\centering
\includegraphics[width=\linewidth]{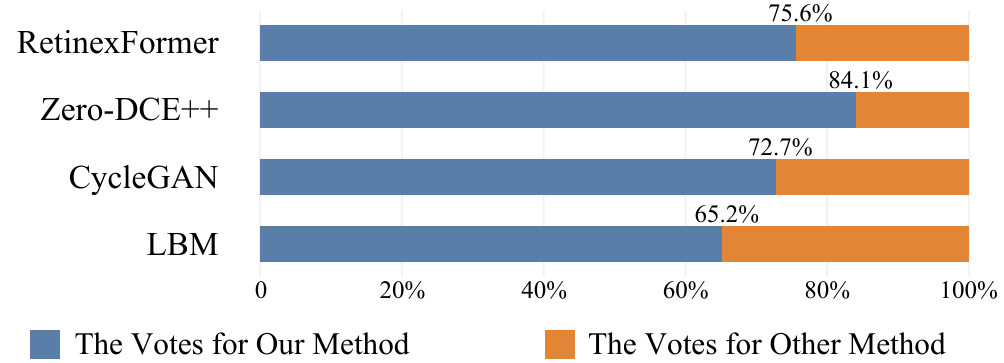}
\caption{User study against representative baselines. Each bar reports the percentage of pairwise votes favoring MrLIE over the compared method, indicating stronger perceived naturalness and restoration quality.}\label{fig-user-study}
\end{figure}

\subsection{Comparison with State-of-the-Art Methods}
We compare MrLIE with two types of pipelines: \textbf{1) supervised methods} (denoted by \textbf{S}), including LLFormer~\cite{wang2023ultra}, RetinexFormer~\cite{cai2023retinexformer}, DFFN~\cite{yao2024spatial}, HVI~\cite{yan2025hvi}, LBM~\cite{chadebec2025lbm}, GT-mean-loss~\cite{liao2025gt}, DATPRL-IR~\cite{dong2026learning}, TPCNet~\cite{shi2026tpcnet}, and SLAIR~\cite{liang2026slair}; and \textbf{2) unsupervised methods} (denoted by \textbf{U}), including Zero-DCE~\cite{guo2020zero}, RUAS~\cite{liu2021retinex}, Zero-DCE++~\cite{li2021learning}, NeRCo~\cite{yang2023implicit}, PairLIE~\cite{fu2023learning}, Zero-IG~\cite{shi2024zero}, AGLLDiff~\cite{lin2025aglldiff}, GEFU~\cite{wang2025enhancement}, and ZeroIDIR~\cite{jiang2026zeroidir}. Notably, we use their officially released models and protocols for fair comparisons.

\textbf{Quantitative Comparison}. Table~\ref{tab:dicm_npe_comparison} and Table~\ref{tab:lime_mef_comparison} report the quantitative results on four real-world ground-view datasets, DICM, NPE, LIME, and MEF. As shown, MrLIE generally achieves the best performance in terms of most metrics against existing methods, even surpassing a state-of-the-art (SOTA) supervised method, GT-mean-loss, by a considerable margin, which lowers BRISQUE from 15.080 (GT-mean-loss) to 8.534 and improves MUSIQ from 58.845 to 69.541 and CLIP-IQA from 0.470 to 0.705 on the DICM dataset. In addition, compared with the recent SOTA unsupervised method ZeroIDIR, our method also demonstrates obviously superior generalizability.

Then, we evaluate MrLIE on real-world nighttime aerial images, which compound low-light degradation with extreme darkness and heterogeneous artificial lights. As illustrated in Table~\ref{tab:quantitative_comparison}, on the unpaired U3D dataset, our method outperforms both types of methods with significant gains. For example, MrLIE exceeds ZeroIDIR by 2.498 in NIQE, 24.351 in MUSIQ, 30.847 in BRISQUE, 1.275 in PI, 0.326 in CLIP-IQA, and  1.79 in LIQE. For iSAID-dark, MrLIE maintains competitive pixel-level scores against recent methods and establishes SOTA performance in perceptual quality.

\textbf{Qualitative Comparison}. Fig.~\ref{fig-dicmnpe} and Fig.~\ref{fig-limenpe} present visual comparisons on DICM, NPE, LIME, and MEF. We can see that existing methods frequently suffer from unbalanced exposure or severe color casts. Specifically, in Fig.~\ref{fig-dicmnpe}, the supervised method SLAIR exhibits severe under-enhancement (1st and 2nd rows). Among the unsupervised methods, RUAS severely overexposes the vehicles and sky (3rd row). GEFU introduces an unnatural green background tint (4th row). Similar limitations also appear on LIME and MEF in Fig.~\ref{fig-limenpe}. GT-mean-loss produces obvious orange artifacts on the helmet (1st row). TPCNet causes cyan color distortion (2nd row). Meanwhile, both RUAS and Zero-IG suffer from severe overexposure and color distortion (3rd and 4th rows). In contrast, MrLIE eliminates these color distortions and restores balanced exposure with faithful local structures across all datasets.


For aerial nighttime scenes on U3D shown in Fig.~\ref{fig-u3d}, existing methods often retain nighttime color casts, blur local geometry, or cause overexposure. For instance, LBM introduces an unnatural greenish tone in the river region in the second row, while Zero-DCE++ causes hazy overexposure in the third row. Empowered by iterative artificial-light suppression during reference synthesis and explicit structural anchoring via COM and structure-anchored injection, MrLIE recovers natural daytime colors while maintaining sharp road networks and building layouts. Consistent advantages are observed on iSAID-dark in Fig.~\ref{fig-iSAID}. DATPRL-IR exhibits insufficient brightening (1st and 2nd rows). GEFU similarly fails to improve visibility in dark regions. By comparison, MrLIE produces the normal-light images with balanced illumination, crisp aerial structures, and natural color appearance.

\textbf{User Study}. We further conduct a user study to evaluate human perceptual preference, as shown in Fig.~\ref{fig-user-study}. We invited 30 participants to perform pairwise comparisons between MrLIE and four representative baselines, including RetinexFormer, Zero-DCE++, CycleGAN, and LBM. For each comparison, participants are asked to select the result with more natural daytime appearance, fewer visible artifacts, and better overall restoration quality. MrLIE obtains the majority of preference votes against all compared methods, with preference ratios of 75.6\%, 84.1\%, 72.7\%, and 65.2\%, respectively. These results indicate that MrLIE is consistently favored by human observers, further validating its perceptual advantage beyond quantitative no-reference metrics.

\subsection{Ablation Study}
We conduct ablation studies on 100 test images from the U3D dataset using no-reference image quality metrics. Specifically, we evaluate the individual contributions of MLLM-guided reference generation, COM, and SAI, alongside the sensitivity to regularization weights.

\begin{table}[!t]
\caption{Ablation study on MLLM-Gen for reference generation (Ref.) and low-light image enhancement (LLIE). One-shot uses a single generation pass; the partial variants omit MLLM-Critic or the IQA scorer from iterative refinement.}
\label{tab:ablation_reference_generation}
\centering
{\setlength{\tabcolsep}{4.5pt}
\footnotesize
\begin{tabular*}{0.8\columnwidth}{@{\extracolsep{\fill}}clccc@{}}
\toprule
\multirow{2}{*}{Type} & \multirow{2}{*}{Strategy} & \multicolumn{3}{c}{Metrics} \\
\cmidrule(lr){3-5}
& & NIQE & BRISQUE & PI \\
\midrule
\multirow{4}{*}{Ref.} & One-shot & 3.406 & \underline{11.826} & 2.324 \\
& w/o MLLM-Critic & \underline{3.203} & 12.481 & \underline{2.210} \\
& w/o IQA Scorer & 3.249 & 13.039 & 2.244 \\
& MLLM-Gen & \textbf{2.957} & \textbf{8.110} & \textbf{2.114} \\
\midrule
\multirow{4}{*}{LLIE}
& One-shot & \textbf{2.851} & 12.837 & 2.322 \\
& w/o MLLM-Critic & 3.008 & 16.246 & 2.516 \\
& w/o IQA Scorer & 2.977 & \underline{9.852} & \underline{2.180} \\
& MLLM-Gen & \underline{2.884} & \textbf{5.368} & \textbf{2.110} \\
\bottomrule
\end{tabular*}}
\end{table}

\begin{figure}[!t]
\centering
\includegraphics[width=\linewidth]{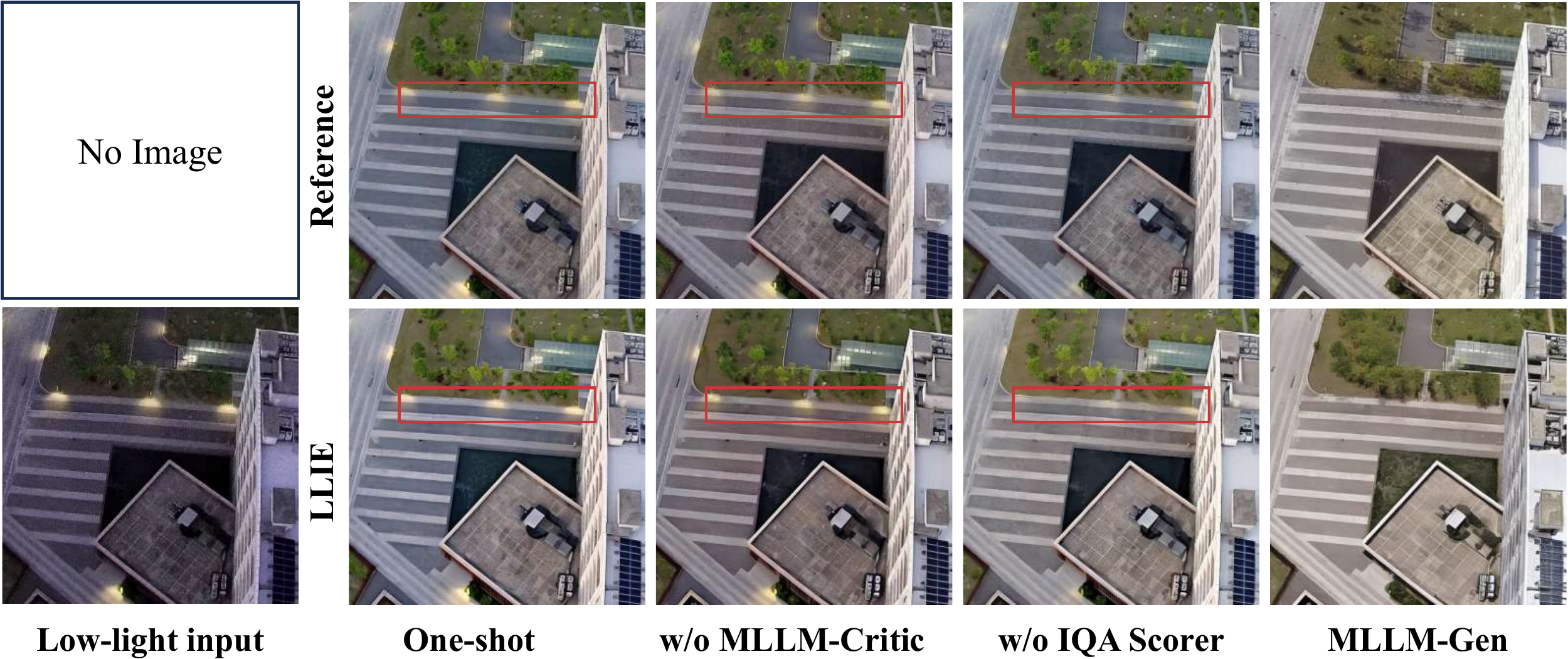}
\caption{Visual ablation of MLLM-guided reference generation. The upper row compares generated references and the lower row shows the corresponding restoration results. Red boxes indicate residual artificial-light cues. The complete MLLM-Gen suppresses nighttime-light residues and provides more reliable daytime guidance for restoration.}\label{fig-ablation-mllm}
\end{figure}

\begin{table}[!t]
\caption{Component ablation of COM. Offset denotes learnable spatial warping of reference features, and Conf denotes confidence-gated fusion for suppressing unreliable reference regions.}
\label{tab:ablation_COM}
\centering
{\setlength{\tabcolsep}{4.5pt}
\footnotesize
\begin{tabular*}{0.8\columnwidth}{@{\extracolsep{\fill}}cccccc@{}}
\toprule
\multirow{2}{*}{Var.} & \multicolumn{2}{c}{Components} & \multicolumn{3}{c}{Metrics} \\
\cmidrule(lr){2-3}\cmidrule(lr){4-6}
 & Conf & Offset & NIQE & BRISQUE & PI \\
\midrule
1 & $\times$ & $\times$ & 3.234 & 7.258 & 2.399 \\
2 & $\times$ & $\checkmark$ & \underline{3.202} & \textbf{4.894} & \underline{2.306} \\
3 & $\checkmark$ & $\checkmark$  & \textbf{2.884} & \underline{5.368} & \textbf{2.110} \\
\bottomrule
\end{tabular*}}
\end{table}

\begin{figure}[!t]
\centering
\includegraphics[width=\linewidth]{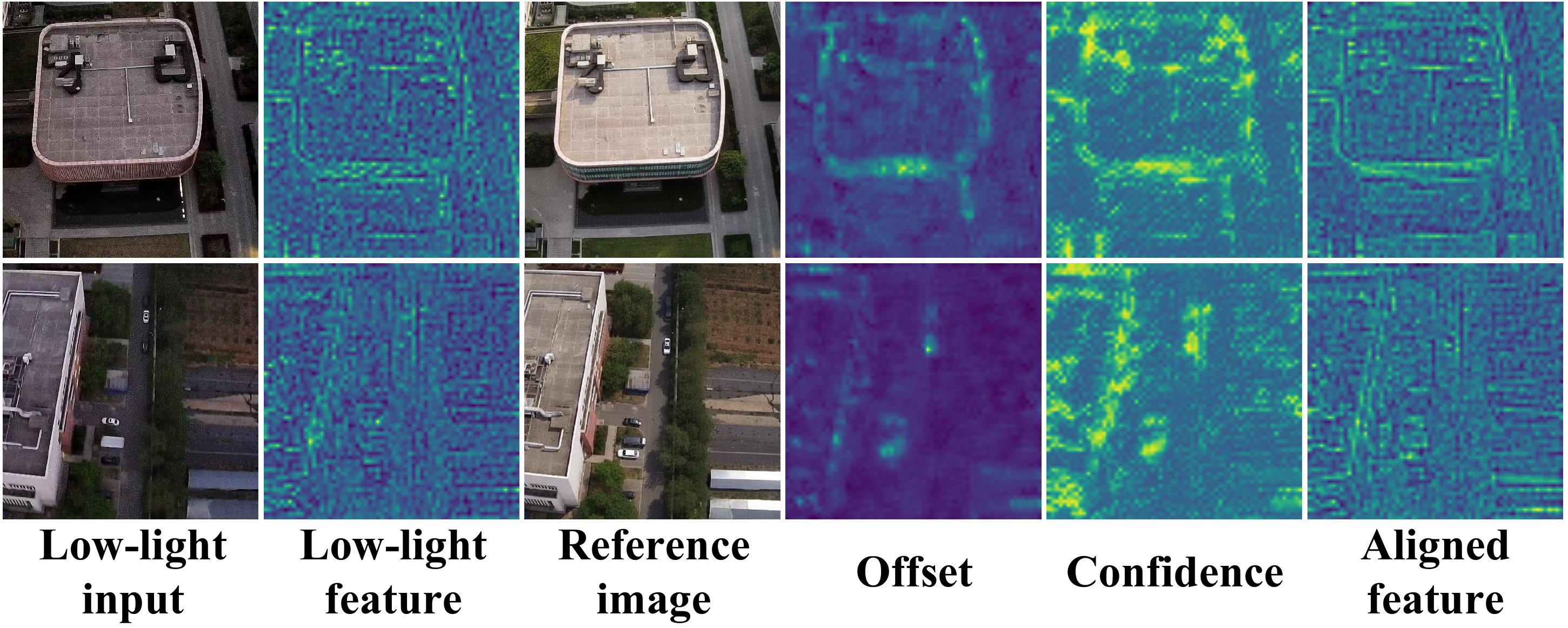}
\caption{Visualization of COM. For each example, we show the low-light input and feature, the generated reference image, the predicted offset field, the confidence map, and the aligned reference feature.}\label{fig-ablation-components}
\end{figure}

\textbf{MLLM-Gen}. The built MLLM-Gen aims to generate normal-light references that serve as guidance to facilitate the subsequent LLIE process. Here, we first examine the generation performance using different strategies (denoted by ``Ref.''). As shown in Table~\ref{tab:ablation_reference_generation}, the complete MLLM-Gen exhibits the most consistent advantage among all strategies. The model using one-shot synthesis can provide promising illumination and spatial structure in BRISQUE but fails in perceptual quality. Using either MLLM-Critic or the IQA scorer alone can lead to biased assessments and less reliable reference generation. By combining all components, MLLM-Gen can yield a more stable reference. Then, we conduct experiments to study the effects of these strategies for LLIE (denoted by ``LLIE''). The full model achieves the best BRISQUE and PI scores, while obtaining the second-best NIQE score. Fig.~\ref{fig-ablation-mllm} provides consistent visual evidence. MLLM-Gen produces an image with plausible textures and natural illumination, whereas others retain visible artificial light residues. This suggests that our method can adaptively absorb the reliable reference prior while correcting unreasonable appearance.

\begin{figure*}[!tbp]
\centering
\includegraphics[width=\textwidth]{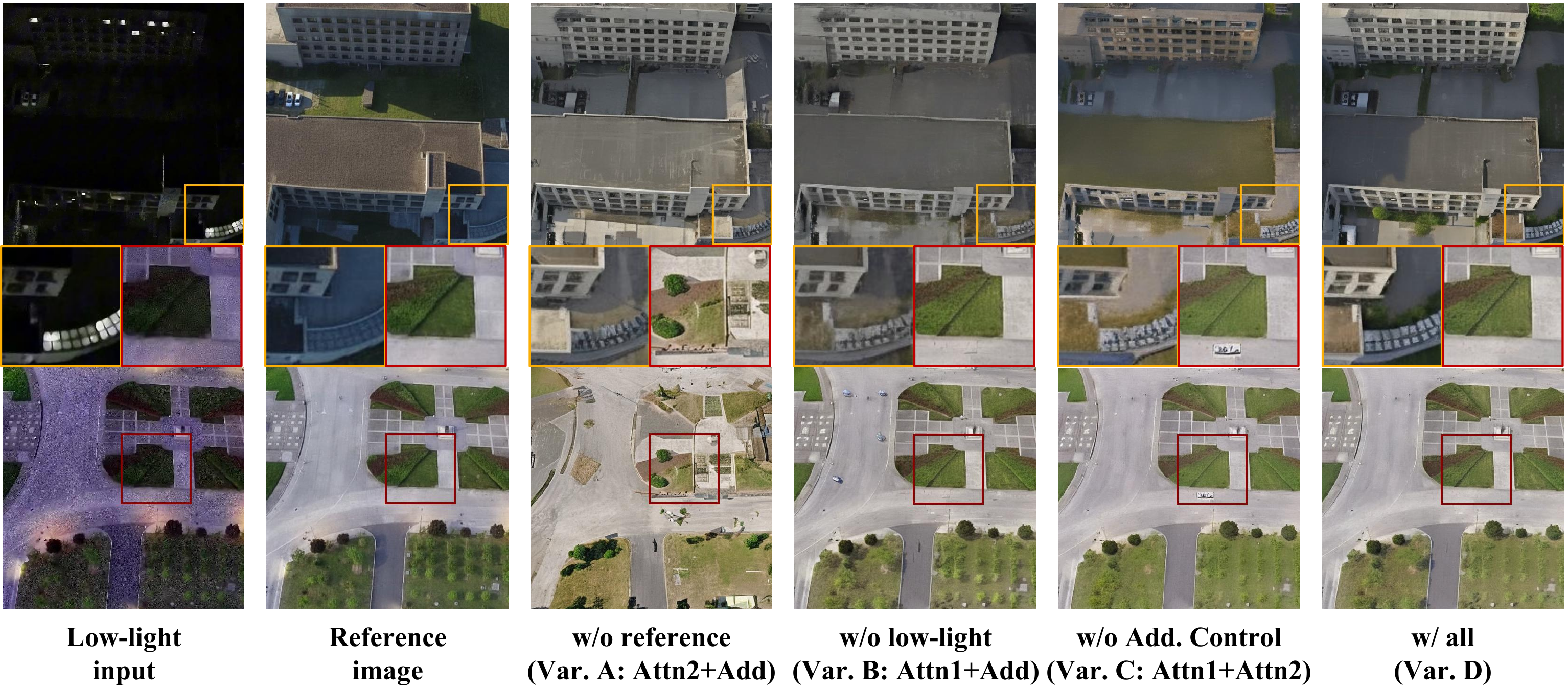}
\caption{Visual ablation of SAI. The variants correspond to Table~\ref{tab:ablation_attn}. Removing Attn1 disables the reference interaction and causes severe structural drift, while removing Attn2 or Add weakens the anchoring to the low-light input. The full model preserves the input layout while recovering more reliable structures.}\label{fig-abla-SAI}
\end{figure*}

\textbf{Effect of COM}. The proposed COM involves offset field estimation and confidence map prediction to correct the spatial misalignment between the low-light input and the reference guidance. In Table~\ref{tab:ablation_COM}, we isolate the two operations. We can see that incorporating the offset yields consistent improvements over the model without both components. Besides, by suppressing unreliable alignment, the confidence branch can continuously improve the overall balance. Fig.~\ref{fig-ablation-components} further visualizes this process, where we can see that the offset field models local reference-input correspondence and the confidence map adaptively gates the warped reference cues according to their reliability, producing aligned features that better follow the low-light structure.

\begin{table}[!t]
\caption{Component ablation of SAI. Attn1 introduces aligned reference cues, Attn2 re-anchors the reference-enhanced representation to the low-light structure, and Add denotes the ControlNet-style additive branch.}
\label{tab:ablation_attn}
\centering
{\setlength{\tabcolsep}{4.5pt}
\footnotesize
\begin{tabular*}{0.8\columnwidth}{@{\extracolsep{\fill}}ccccccc@{}}
\toprule
\multirow{2}{*}{Var.} & \multicolumn{3}{c}{Components} & \multicolumn{3}{c}{Metrics} \\
\cmidrule(lr){2-4}\cmidrule(lr){5-7}
 & Attn1 & Attn2 & Add & NIQE & BRISQUE & PI \\
\midrule
A & $\times$ & $\checkmark$ & $\checkmark$ & \underline{3.017} & 10.196 & 2.281 \\
B & $\checkmark$ & $\times$ & $\checkmark$ & 3.073 & 7.770 & 2.311 \\
C & $\checkmark$ & $\checkmark$ & $\times$ & 3.097 & \underline{5.398} & \underline{2.233} \\
D & $\checkmark$ & $\checkmark$ & $\checkmark$ & \textbf{2.884} & \textbf{5.368} & \textbf{2.110} \\
\bottomrule
\end{tabular*}}
\end{table}

\textbf{Effect of SAI}. In SAI, the first cross-attention module (Attn1) uses aligned reference control as the key and value to provide normal-light appearance. The second cross-attention (Attn2) applies low-light features as the key and value to constrain structural semantics. In addition, a standard additive pathway with zero convolution is used to inject the ControlNet signal, providing explicit structural anchoring (Add.). As shown in Table~\ref{tab:ablation_attn}, when all three branches are enabled, the full SAI model achieves the best performance across all metrics, validating its effectiveness.

To more intuitively illustrate the influence of the three components above, in Fig.~\ref{fig-abla-SAI}, we visualize the corresponding results. Once the reference is removed, the model (Var. A) exhibits clear structural instability, producing a road boundary whose orientation is inconsistent with the low-light input. The model also suffers from large-scale content confusion. Removing Attn2 weakens the constraint that pulls reference-enhanced features back to the input structure, while removing Add discards the explicit ControlNet-style structural pathway. Both variants reduce the structural anchoring effect.

%


\begin{table}[!t]
\caption{Sensitivity analysis of the COM regularization weights. The two blocks sweep the confidence and offset regularizers around the default setting.}
\label{tab:ablation_sga_regularization}
\centering
{\setlength{\tabcolsep}{4.5pt}
\footnotesize
\begin{tabular*}{0.8\columnwidth}{@{\extracolsep{\fill}}ccccc@{}}
\toprule
\multicolumn{2}{c}{Weights} & \multicolumn{3}{c}{Metrics} \\
\cmidrule(lr){1-2}\cmidrule(lr){3-5}
$\lambda_{\mathrm{off}}$ & $\lambda_{\mathrm{conf}}$ & NIQE & BRISQUE & PI \\
\midrule
0 & 0 & \underline{2.995} & 5.483 & \underline{2.168}\\
0.001 & 0 & 3.069 & 6.815 & 2.373 \\
0.001 & 0.0001 & 3.204 & \textbf{4.606} & 2.275\\
0.001 & 0.0005 & \textbf{2.884} & 5.368 & \textbf{2.110} \\
0.001 & 0.001 & 3.244 & \underline{5.066} & 2.322 \\
\midrule
0 & 0.0005 & \underline{2.898} & 7.801 & 2.282 \\
0.0005 & 0.0005 & 3.184 & \textbf{3.356} & \underline{2.261} \\
0.001 & 0.0005 & \textbf{2.884} & \underline{5.368} & \textbf{2.110} \\
0.005 & 0.0005 & 2.983 & 7.525 & 2.329 \\
\bottomrule
\end{tabular*}}
\end{table}

\begin{figure*}[!tbp]
\centering
\includegraphics[width=\textwidth]{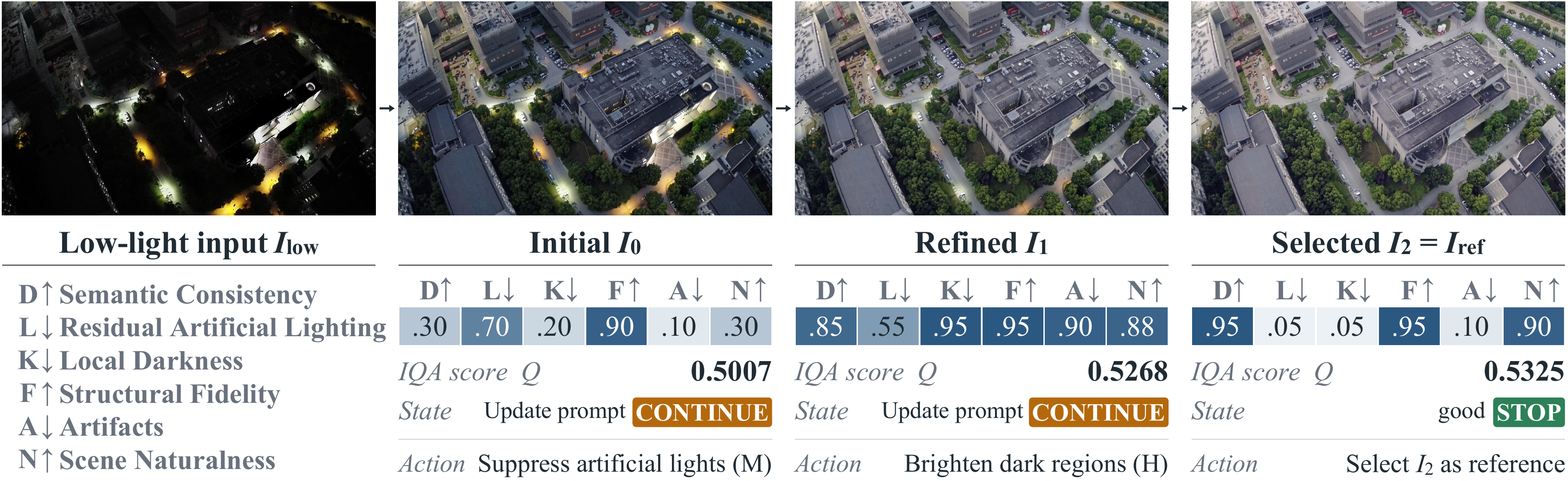}
\caption{Iterative refinement of a daytime reference through diagnosis-guided prompt updates. Each candidate is accompanied by diagnostic scores and the resulting action, illustrating how artificial-light suppression and dark-region brightening lead to the selected reference.}\label{fig-iter}
\end{figure*}

\textbf{Loss Functions}. In this work, besides the typical diffusion denoising loss function, the training for MrLIE also contains an additional offset smoothness loss (Eq. (13)) and a confidence regularization loss (Eq. (14)). As reported in Table~\ref{tab:ablation_sga_regularization}, removing the offset regularization can make the alignment field less stable, while removing the confidence regularization may cause the model to over-trust or under-use the generated reference. The setting $\lambda_{\mathrm{off}}=10^{-3}$ and $\lambda_{\mathrm{conf}}=5\times10^{-4}$ achieves the best NIQE and PI in the main sweep, showing a good balance between flexible reference alignment and stable reference gating.

\subsection{Analysis of Iterative Refinement}
Fig.~\ref{fig-iter} illustrates the diagnosis-driven closed loop of Sec.~\ref{sec:iterative_ref_generation} on a representative nighttime aerial scene. Given a low-light input $I_{\mathrm{low}}$ containing shadows and cluttered vehicle lights, MLLM-Gen generates an initial candidate $I_0$ under the base prompt $p_0$. Structural fidelity is high, but single-step generation leaves artificial glare on the ground. This gives a relatively low overall score $Q$ and a high artificial-light residual score $L$. The decision policy therefore continues iterating and first triggers the ``suppress artificial light'' action. In the revised candidate $I_1$, street-lamp glare is clearly weakened, $L$ drops, and $Q$ rises. The policy then triggers the ``brighten dark regions'' action to reduce the dark-region residual score $K$, thereby removing the remaining shadows. The resulting candidate $I_2$ clears both the residual illumination and the local dark areas, lowers $L$ and $K$ to $0.05$, and raises $Q$ to its peak of $0.5325$. This satisfies the termination criterion, so $I_2$ is selected as the final reference image $I_{\mathrm{ref}}$. The results show that the multimodal closed-loop diagnosis can autonomously correct generation defects and provide a reliable daytime prior for the subsequent restoration.

\section{Conclusion}
In this paper, we introduce MrLIE, a novel MLLM-guided aligned-reference diffusion framework for real-world LLIE. MrLIE develops an iterative MLLM-guided reference generation loop that leverages semantic diagnosis and perceptual assessment to generate normal-light references that provide structural semantics and high-quality appearance. To ensure spatial consistency, a COM is proposed to rectify the misalignment, where these aligned cues are subsequently integrated into a diffusion backbone through a structure-anchored injection mechanism, synergizing reference semantics with low-light structural constraints. Extensive evaluations on various benchmarks demonstrate that MrLIE achieves superior perceptual quality and robust generalization across diverse scenarios.


\bibliographystyle{IEEEtran}
\bibliography{IEEEabrv, paper}

\end{document}